\documentclass[10pt,twocolumn,letterpaper]{article}

\usepackage[pagenumbers]{cvpr} 

\usepackage{graphicx}
\usepackage{amsmath}
\usepackage{amssymb}
\usepackage{booktabs}
\usepackage{setspace}
\usepackage{enumitem}
\setlist{leftmargin=3.5mm}

\usepackage[table]{xcolor}
\definecolor{LightCyan}{rgb}{0.88,1,1}
\definecolor{Gray}{gray}{0.92}
\definecolor{LightGray}{gray}{0.9486}

\usepackage[pagebackref,breaklinks,colorlinks]{hyperref}

\usepackage[capitalize]{cleveref}
\crefname{section}{Sec.}{Secs.}
\Crefname{section}{Section}{Sections}
\Crefname{table}{Table}{Tables}
\crefname{table}{Tab.}{Tabs.}

\def\cvprPaperID{2800} 
\def\confName{CVPR}
\def\confYear{2023}

\begin{document}

\title{Towards Efficient Use of Multi-Scale Features in Transformer-Based\\Object Detectors}

\author{
\fontsize{11.2}{13.5}\selectfont
Gongjie Zhang$^\dagger$$^{1,2}$\quad Zhipeng Luo$^\dagger$$^{1,3}$\quad Zichen Tian$^1$\quad Jingyi Zhang$^1$\quad Xiaoqin Zhang$^4$\quad Shijian Lu\thanks{\,marks corresponding author.\;\;\; $^\dagger$ marks equal technical contribution.}\,\,$^{1}$ \smallskip\\
{\fontsize{9.15}{11.5}\selectfont $^{1}$S-Lab,\;Nanyang\,Technological\,University\quad $^{2}$Black\,Sesame\,Technologies\quad $^{3}$SenseTime\,Research\quad $^{4}$Wenzhou\,University} \\
{\tt\footnotesize gjz@ieee.org \qquad zhipeng001@e.ntu.edu.sg \qquad shijian.lu@ntu.edu.sg}
}

\maketitle

\begin{abstract}

Multi-scale features have been proven highly effective for object detection but often come with huge and even prohibitive extra computation costs, especially for the recent Transformer-based detectors. In this paper, we propose Iterative Multi-scale Feature Aggregation (IMFA) -- a generic paradigm that enables efficient use of multi-scale features in Transformer-based object detectors. The core idea is to exploit sparse multi-scale features from just a few crucial locations, and it is achieved with two novel designs. First, IMFA rearranges the Transformer encoder-decoder pipeline so that the encoded features can be iteratively updated based on the detection predictions. Second, IMFA sparsely samples scale-adaptive features for refined detection from just a few keypoint locations under the guidance of prior detection predictions. As a result, the sampled multi-scale features are sparse yet still highly beneficial for object detection. Extensive experiments show that the proposed IMFA boosts the performance of multiple Transformer-based object detectors significantly yet with only slight computational overhead.

\end{abstract}

\let\oldfootnote\thefootnote
\renewcommand{\thefootnote}{}

\section{Introduction}     \label{sec:IMFA_introduction}

Detecting objects of vastly different scales has always been a major challenge in object detection~\cite{Liu2019DeepLF}. Fortunately, strong evidence~\cite{FPN,tridentnet,efficientdet,graphfpn,SMCA-DETR,DeformableDETR} shows that object detectors can significantly benefit from multi-scale features while dealing with large scale variation. For ConvNet-based object detectors like Faster R-CNN~\cite{FasterRCNN} and FCOS~\cite{FCOS}, Feature Pyramid Network (FPN)~\cite{FPN} and its variants~\cite{parallelfpn,PANet,m2det,PanopticFPN,NASFPN,efficientdet,graphfpn} have become the go-to components for exploiting multi-scale features. \footnotetext{Project Page: \href{https://github.com/ZhangGongjie/IMFA}{https://github.com/ZhangGongjie/IMFA}{\tiny\,}.}

\begin{figure}[t]
\begin{center}
\includegraphics[width=0.82\linewidth]{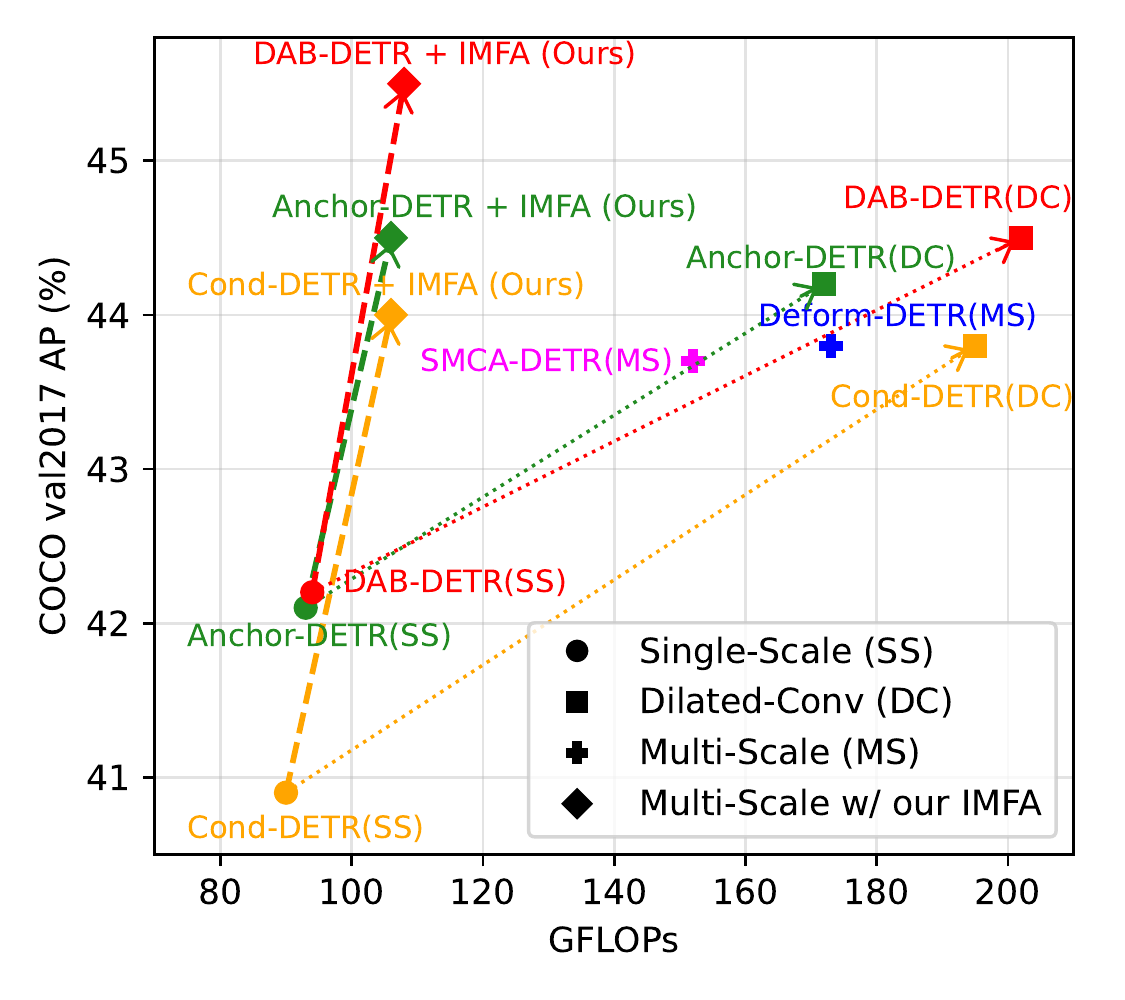}
\end{center}
\vspace{-5.0mm}
\caption{The proposed \textit{Iterative Multi-scale Feature Aggregation (IMFA)} is a generic approach for efficient use of multi-scale features in Transformer-based object detectors. It boosts detection accuracy on multiple object detectors at minimal costs of additional computational overhead. Results are obtained with ResNet-50. Best viewed in color.}\label{fig:AP_GFLOPs}
\vspace{-1.8mm}
\end{figure} 

Other than ConvNet-based object detectors, the recently proposed DEtection TRansformer (DETR)~\cite{DETR} has established a fully end-to-end object detection paradigm with promising performance. However, naively incorporating multi-scale features using FPN in these Transformer-based detectors~\cite{DETR,DeformableDETR,SMCA-DETR,ConditionalDETR,AnchorDETR,DABDETR,SAM-DETR,DN-DETR} often brings enormous and even unfeasible computation costs, primarily due to the poor efficiency of the attention mechanism in processing high-resolution features. Concretely, to handle a feature map with a spatial size of $H \times W$, ConvNet requires a computational cost of $O(HW)$, while the complexity of the attention mechanism in Transformer-based object detectors is $O(H^{2}W^{2})$. To mitigate this issue, Deformable DETR~\cite{DeformableDETR} and Sparse DETR~\cite{SparseDETR} replace the original global dense attention with sparse attention. SMCA-DETR~\cite{SMCA-DETR} restricts most Transformer encoder layers to be scale-specific, with only one encoder layer to integrate multi-scale features. However, as the number of tokens increases quadratically\,\textit{w.r.t.}\,feature map size (typically 20x$\sim$80x compared to single-scale), these methods are still costly in computation and memory consumption, and rely on special operators like deformable attention~\cite{DeformableDETR} that introduces extra complexity for deployment. To the best of our knowledge, there is yet no generic approach that can efficiently exploit multi-scale features for Transformer-based object detectors.

In this paper, we present \textit{Iterative Multi-scale Feature Aggregation (IMFA)}, a concise and effective technique that can serve as a generic paradigm for efficient use of multi-scale features in Transformer-based object detectors. The motivation comes from two key observations: \textit{(i)} the computation of high-resolution features is highly redundant as the background usually occupies most of the image space, thus only a small portion of high-resolution features are useful to object detection; \textit{(ii)} unlike ConvNet, the Transformer's attention mechanism does not require grid-shaped feature maps, which offers the feasibility of aggregating multi-scale features only from some specific regions that are likely to contain objects of interest. The two observations motivate us to sparsely sample multi-scale features from just a few informative locations and then aggregate them with encoded image features in an iterative manner.

Concretely, IMFA consists of two novel designs in the Transformer-based detection pipelines.
\textit{First}, IMFA rearranges the encoder-decoder pipeline so that each encoder layer is immediately connected to its corresponding decoder layer. This design enables iterative update of encoded image features along with refined detection predictions.
\textit{Second}, IMFA sparsely samples multi-scale features from the feature pyramid generated by the backbone, with the sampling process guided by previous detection predictions. Specifically, motivated by the spatial redundancy of high-resolution features, IMFA only focuses on a few promising regions with high likelihood of object occurrence based on prior predictions. Furthermore, inspired by the significance of objects' keypoints for recognition and localization~\cite{ExtremeNet,reppoints,SAM-DETR,qiu2020borderdet}, IMFA first searches several keypoints within each promising region, and then samples useful features around these keypoints at adaptively selected scales. The sampled features are finally fed to the subsequent encoder layer along with the image features encoded by the previous layer. With the two new designs, the proposed IMFA aggregates only the most crucial multi-scale features from those informative locations. Since the number of the aggregated features is small, IMFA introduces minimal computational overhead while consistently improving the detection performance of Transformer-based object detectors. It is noteworthy that IMFA is a generic paradigm for efficient use of multi-scale features: \textit{(i)} as shown in Fig.~\ref{fig:AP_GFLOPs}, it can be easily integrated with multiple Transformer-based object detectors with consistent performance boosts; \textit{(ii)} as discussed in Section~\ref{sec:IMFA_exp_kpt_det}, IMFA has the potential to boost DETR-like models on tasks beyond object detection.

\vspace{+2.3mm}
\noindent
To summarize, the contributions of this work are threefold.
\vspace{-2.66mm}
\begin{itemize}
\setlength\itemsep{-0.8mm}
  \item  We propose a novel DETR-based detection pipeline, where encoded features can be iteratively updated along with refined detection predictions. This new pipeline allows to leverage intermediate predictions as guidance for robust and efficient multi-scale feature encoding.
  \item We propose a sparse sampling strategy for multi-scale features, which first identifies several promising regions under the guidance of prior detections, then searches several keypoints within each promising region, and finally samples their features at adaptively selected scales. We demonstrate that such sparse multi-scale features can significantly benefit object detection.
  \item Based on the two contributions above, we propose \textit{Iterative Multi-scale Feature Aggregation (IMFA)} -- a simple and generic paradigm that enables efficient use of multi-scale features in Transformer-based object detectors. IMFA consistently boosts detection performance on multiple object detectors, yet remains computationally efficient. This is the pioneering work that investigates a generic approach for exploiting multi-scale features efficiently in Transformer-based object detectors.
\end{itemize}

\begin{figure*}[t!] 
\begin{center}
   \vspace{-1.0mm}
   \includegraphics[width=0.90\linewidth]{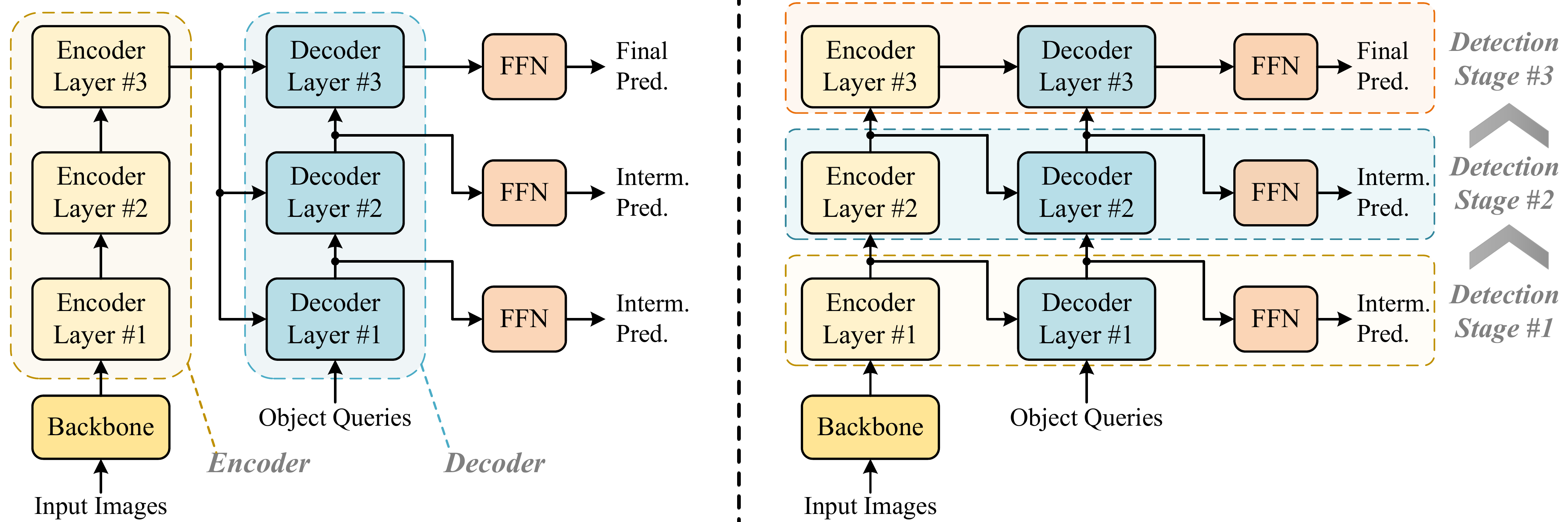}
\end{center}
\vspace{-5mm}
   \caption{
   \textit{\textbf{Left:}} Most existing Transformer-based object detectors employ stacked Transformer encoder layers to obtain a fixed set of encoded image features, which are fed to each Transformer decoder layer to interact with object queries. Only object queries and their corresponding detection predictions are iteratively updated. \textit{\textbf{\;\;Right:}} IMFA rearranges the Transformer encoder-decoder pipeline into multiple stacked \textit{detection stages}. Each \textit{detection stage} is composed of an encoder layer, a decoder layer, and a feed-forward network (FFN), in which encoded features, object queries, and detection predictions can all be iteratively updated during the detection refinement process. Only three encoder and decoder layers are presented for concise illustration.
   }
\label{fig:difference}
\vspace{-0.25mm}
\end{figure*}

\section{Related Work}
\label{sec:related_work}

\noindent
\textbf{Object Detection.\;}
Most modern object detectors, like Faster R-CNN~\cite{FasterRCNN}, YOLO~\cite{YOLO9000}, and FCOS~\cite{FCOS}, are ConvNet-based. They have achieved promising results on various detection benchmarks\cite{CascadeRCNN,CADNet,LibraRCNN,masktextspotter,efficientdet,StrongWeak,FewshotReweighting,PNPDet,fsod,fsdet,FSDetView}. However, these methods detect objects by defining surrogate regression and classification tasks, which rely on many hand-crafted components, such as anchors, rule-based training target assignment, and non-maximum suppression (NMS). Thus the detection pipelines of these ConvNet-based detectors are complex, hyper-parameter-intensive, and not fully end-to-end, leading to sub-optimal performance. Unlike ConvNet-based detectors, the recently proposed DETR~\cite{DETR} has revolutionized the paradigm for object detection using a Transformer~\cite{transformer} encoder-decoder architecture, eliminating the need for those hand-crafted components. Inspired by DETR~\cite{DETR}, many Transformer-based object detectors~\cite{PnPDETR,DeformableDETR,up-detr,MDETR,YOLOS,CF-DETR,DN-DETR,3DETR,DABDETR,SAM-DETR,SAM-DETR-PlusPlus,OW-DETR,detreg,ViDT,FP-DETR,Meta-DETR,DINO,MapTR,TransPillars,DA-DETR} are proposed and achieve state-of-the-art detection accuracy as well as fast convergence.

\vspace{+1.5mm}
\noindent
\textbf{Multi-Scale Features for Object Detection.\;}
One major challenge in object detection is to effectively represent objects at distinct scales. This is especially crucial for detecting small objects in images. In modern ConvNet-based detectors~\cite{FasterRCNN,focalloss,m2det,FCOS,efficientdet,fsdet,MPSR}, Feature Pyramid Network (FPN)~\cite{FPN} and its variants~\cite{parallelfpn,PANet,m2det,NASFPN,graphfpn} have become the go-to solutions to exploit multi-scale features. However, as feature pyramids require computation on high-resolution feature maps, FPN and its variants also introduce substantial computational overhead.

Multi-scale features are also helpful for Transformer-based object detectors. However, due to the inefficiency of Transformer's attention mechanism~\cite{transformer} to process high-resolution feature maps, it requires special modifications to reduce the computational complexity to a feasible level. Concretely, Deformable DETR~\cite{DeformableDETR} proposes deformable attention, which reduces the complexity via key sparsification in the attention module. SMCA-DETR~\cite{SMCA-DETR} uses only one multi-scale attention encoder layer while restricting other layers to be scale-specific. CF-DETR~\cite{CF-DETR} embeds the Transformer encoder into an FPN~\cite{FPN} to produce feature pyramids, and extracts multi-scale features with RoIAlign~\cite{MaskRCNN}. These methods enable the use of multi-scale features in Transformer-based detectors, but introduce huge computational overhead, require large-memory GPUs for training and inference, and rely on special operators like deformable attention or RoIAlign. To the best of our knowledge, there is yet no generic approach to efficiently leverage multi-scale features for Transformer-based detectors so far.

\vspace{+0.2mm}
\noindent
\textbf{Spatial Redundancy and Sparse Features.\;}
Not all features are equally important. In most cases, only a small portion of features are crucial for recognition. With this motivation, several works~\cite{perforatedCNNs,figurnov2017spatially,SBNet,verelst2020dynamic,PnPDETR,DeformableDETR,SparseDETR} perform sparse operations over feature maps to avoid computation at less informative locations. Specifically, in object detection, AutoFocus~\cite{autofocus} first predicts and crops regions at coarse scales, and then makes final predictions on those regions at a higher resolution. PnP-DETR~\cite{PnPDETR} and Sparse DETR~\cite{SparseDETR} adaptively allocate encoding operations to informative feature tokens. One similar work to our proposed IMFA is QueryDet~\cite{QueryDet}, which first coarsely predicts over low-resolution features, and then sparsely exploits multi-scale features based on the coarse predictions to generate the final detection results, thus improving inference speed. However, unlike our proposed IMFA, QueryDet is designed for single-stage ConvNet-based detectors with FPN~\cite{FPN}, and it only accelerates the inference procedure.

Our proposed IMFA is also inspired by the spatial redundancy in high-resolution features. IMFA only exploits sparse features from only a few highly informative locations to get the best of both worlds for Transformer-based detectors -- high detection accuracy and low computational cost.

\begin{figure*}[t] 
\begin{center}
\vspace{-0.88mm}
   \includegraphics[width=0.91\linewidth]{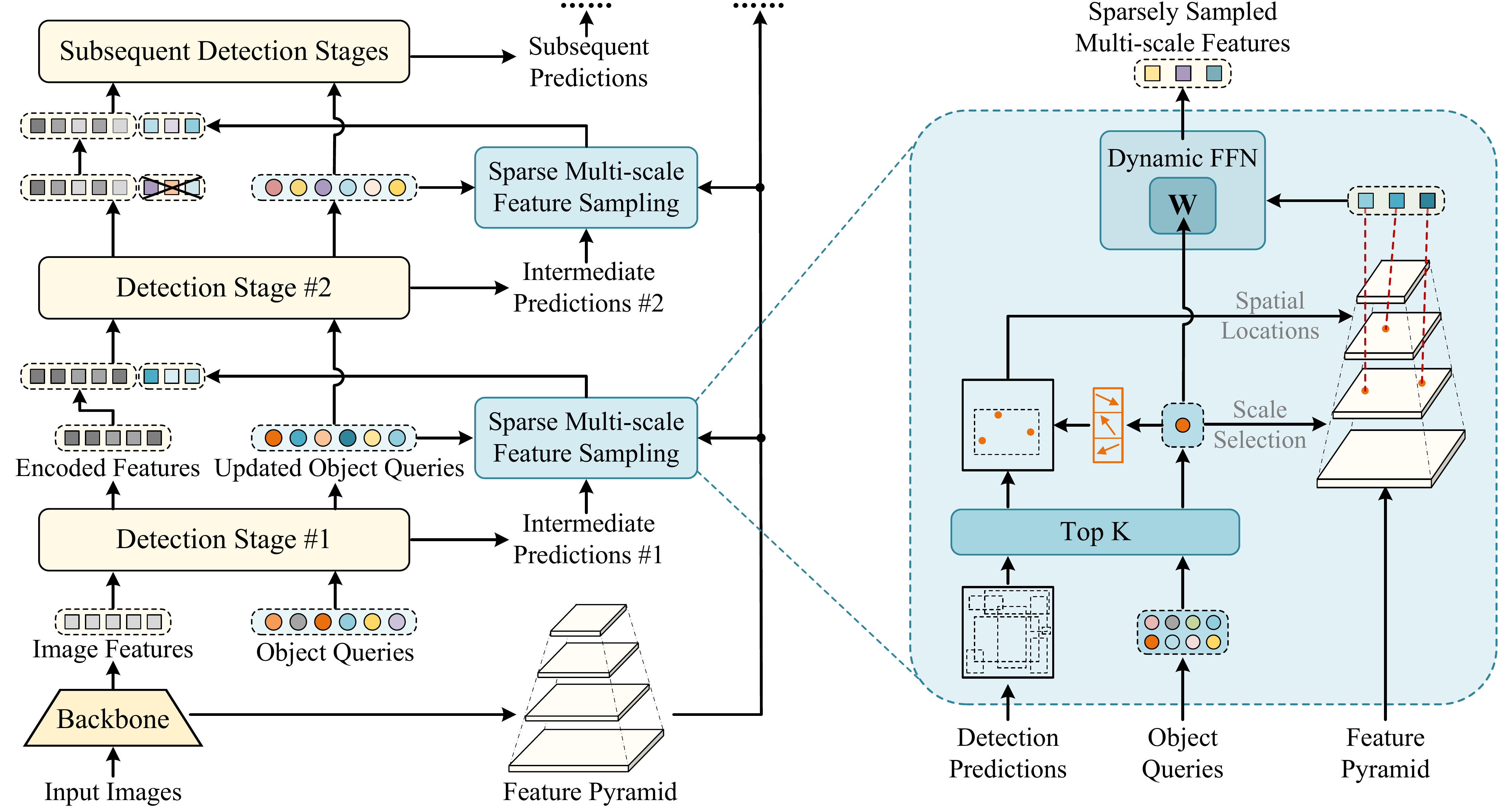}
\end{center}
\vspace{-5.20mm}
   \caption{
   \textbf{The detection pipeline of \textit{Iterative Multi-scale Feature Aggregation (IMFA)}}. IMFA adopts the pipeline in Fig.\,\ref{fig:difference}\,(right) with multiple stacked detection stages, which enables the iterative update of encoded features. On this basis, IMFA performs sparse multi-scale feature sampling under the guidance of prior detection predictions. Specifically, it only focuses on a few promising regions guided by prior detection predictions, then searches for several keypoints within each promising region, and finally samples features around these keypoints at adaptively selected scales. IMFA also adopts a Dynamic FFN to enhance the representation capacity of sparsely sampled multi-scale features by incorporating semantics from their corresponding object queries. The sampled features are fed into the subsequent detection stages along with encoded features for refined detection. Only the first two detection stages are presented for concise illustration.
   }
\label{fig:IMFA}
\end{figure*}

\section{A Revisit of Transformer-Based Detection}
\label{sec:IMFA_revisit}

Since our proposed method is developed on top of the recently proposed Transformer-based object detectors, we first briefly review the detection pipeline of Transformer-based object detectors~\cite{DETR,ConditionalDETR,AnchorDETR,DABDETR}, taking the pioneering work DETR~\cite{DETR} as an example.

DETR~\cite{DETR} formulates object detection as a direct set prediction problem and uses a Transformer~\cite{transformer} encoder-decoder architecture to solve it. Given an image $\mathbf{I} \in \mathbb{R}^{H_{0} \times {W_{0}} \times 3}$, the backbone network generates its feature maps, which are further fed to the Transformer encoder to produce the encoded image features $\mathbf{F} \in \mathbb{R}^{HW \times d}$, where $d$ denotes the feature dimension, and $H_{0}$, $W_{0}$ and $H$, $W$ are the spatial sizes of the input image and its feature maps, respectively. Then, the encoded features are fed to the Transformer decoder to interact with a set of object queries representing potential objects at different spatial locations. The object queries are finally used to produce final detection predictions with a feed-forward network (FFN). The entire detection pipeline is supervised by a set-based global loss with bipartite matching.

Specifically, both the Transformer encoder and decoder are composed of multiple layers. As shown in Fig.~\ref{fig:difference}\,(left), existing methods~\cite{DETR,DeformableDETR,ConditionalDETR,SMCA-DETR,AnchorDETR,DABDETR} usually process the input image features with a stack of encoder layers and obtain a fixed set of encoded features, which are further fed to the Transformer decoder layers to update the detection results iteratively. Differently, as illustrated in Fig.~\ref{fig:difference}\,(right), one major difference introduced by IMFA is that it rearranges the encoder-decoder pipeline into multiple stacked detection stages, so that encoded features can also be iteratively updated along with refined detection predictions. This design modification lays the foundation for efficient use of multi-scale features guided by prior detection results, which is to be detailed in the next section.

\section{Iterative Multi-Scale Feature Aggregation} \label{sec:IMFA_method}

\subsection{Overview}    \label{sec:IMFA_method_overview}

\textit{Iterative Multi-scale Feature Aggregation (IMFA)} is a generic paradigm for efficient use of multi-scale features in Transformer-based object detectors, such as DETR~\cite{DETR}. Fig.~\ref{fig:IMFA} illustrates the detection pipeline of the proposed IMFA. For computational efficiency, IMFA exploits multi-scale features with dual-sparsity: \textit{(i)} it samples multi-scale features from just a few promising regions with high likelihood of object occurrence as guided by prior detection predictions; \textit{(ii)} for each promising region, it only samples features from several keypoints with the most informative features at adaptively selected scales. The dual-sparsity is achieved with two novel designs, which are to be described in detail in the following subsections.

\begin{figure*}[t!] 
\begin{center}
\vspace{-1.8mm}
\includegraphics[width=0.915\linewidth]{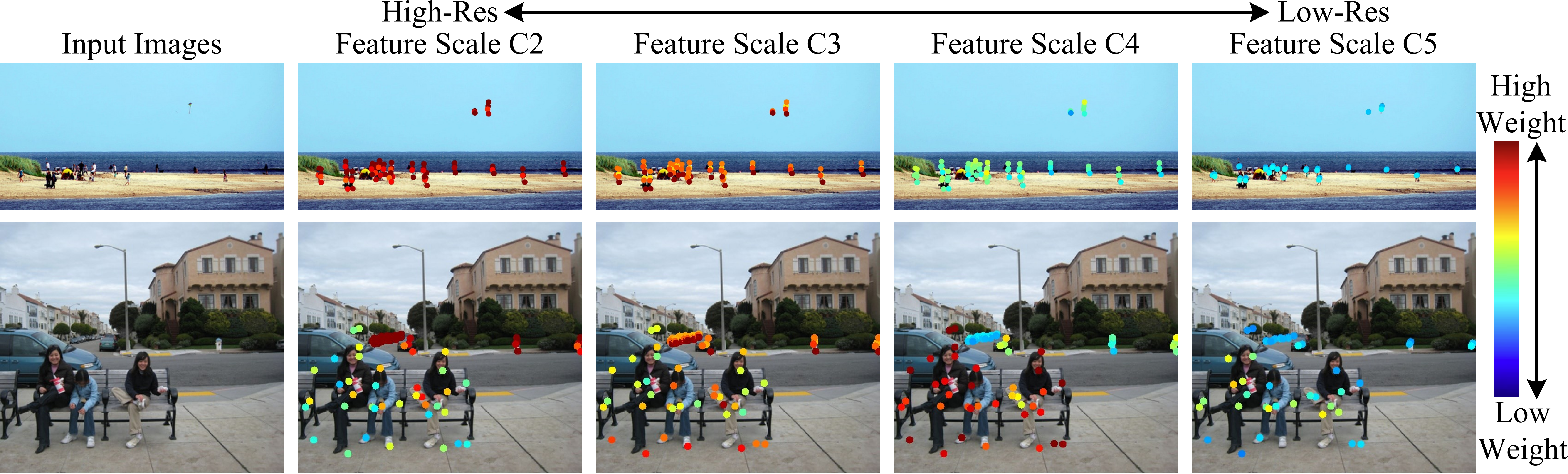}
\end{center}
\vspace{-5mm}
   \caption{
   \textbf{Visualization of IMFA's sampling locations and their adaptively selected feature scales.} The searched sampling points mostly fall around the objects of interest, many of which are highly representative points with rich semantics, such as objects' extremities. Besides, IMFA adaptively selects appropriate feature scales for each sampling point, generating sparse yet informative scale-adaptive features for refined detection predictions. Best viewed in color. More visualizations are provided in technical appendix. 
   }
\vspace{-0.666mm}
\label{fig:IMFA_vis}
\end{figure*}

\subsection{Iterative Update of Encoded Features}  \label{sec:IMFA_method_iter_update}

The iterative update of encoded image features is the basis for IMFA to exploit multi-scale features efficiently. As introduced in Section~\ref{sec:IMFA_revisit}, most existing Transformer-based detectors use fixed encoded image features to make predictions. In order to guide the multi-scale sampling process with prior detections, IMFA rearranges the Transformer encoder-decoder pipeline, as shown in Fig.~\ref{fig:difference}\,(right).

Specifically, instead of using stacked encoder layers to produce a fixed set of feature tokens at one go, IMFA rearranges the detection pipeline into several stacked \textit{detection stages}. Each \textit{detection stage} consists of an encoder layer, a decoder layer, and an FFN. This design lays the foundation for incorporating sparse multi-scale features dynamically under the guidance of prior detection predictions, which is detailed in Section~\ref{sec:IMFA_method_sparse}. It is noteworthy that, according to the experiments in Section~\ref{sec:IMFA_exp_ablation}, this design alone (shown in Fig.\,\ref{fig:difference}\,(right), without incorporating multi-scale features) brings no performance gain over the baseline model.

\subsection{Sparse Feature Sampling and Aggregation}     \label{sec:IMFA_method_sparse}

Naively incorporating multi-scale features into the encoder leads to prohibitive computational complexity, as the number of feature tokens from all scales is too large to be processed by the attention mechanism. This motivates us to exploit only the most informative multi-scale features.

On the basis of Section\;\ref{sec:IMFA_method_iter_update}, IMFA further performs sparse multi-scale feature sampling using prior detection predictions as guidance, as illustrated in Fig.\,\ref{fig:IMFA}. Specifically, IMFA first identifies a few promising regions with high likelihood of object occurrence. Then, it searches for several representative and informative keypoints within each promising region and samples their features at adaptively selected scales. Finally, the sampled features are fed to the subsequent encoder layers to aggregate with single-scale image features to produce refined detection predictions.

\vspace{+1.2mm}
\noindent
\textbf{Identifying Promising Regions Based on Prior Predictions.\;}
In most cases, objects are sparsely distributed across images~\cite{MSCOCO,autofocus,QueryDet}, which motivates us to exploit only the multi-scale features related to these objects. An intuitive solution is to guide the sampling process with the high-confidence detection predictions from the previous detection stage. Concretely, as shown in Fig.~\ref{fig:IMFA}, for each detection stage except the first stage, we select $K$ predictions with the highest classification confidence scores from the previous detection stage as the promising regions. Here, $K = N \times r$, with $N$ denoting the number of object queries and $r$ denoting IMFA's sampling ratio.
Formally, we denote the selected box predictions and their corresponding object queries as $\{(\mathbf{B}_1, \mathbf{Q}_1), ..., (\mathbf{B}_K, \mathbf{Q}_K)\}$.
The multi-scale features are then sampled within these promising regions, which is to be introduced in detail later. Since Transformer-based object detectors~\cite{DETR,ConditionalDETR,SMCA-DETR,AnchorDETR,DABDETR} already employ a sparse set (typically 100$\sim$300) of object queries to represent different objects, the promising regions sampled by IMFA remain sparse for efficient computation.

\begin{table*}[t!]
\begin{center}
\centering
\vspace{-1.66mm}
\setlength{\tabcolsep}{3.8888pt}
\resizebox{0.925\textwidth}{!}{
\begin{tabular}[t]{l|c|ccccc|cccccc}

\toprule[1.5pt]
Method & High-Res\,Feat & \#Epochs & \#Params & FLOPs & FPS & GPU\,Mem & AP & AP$_{\rm 50}$ & AP$_{\rm 75}$ & AP$_{\rm S}$ & AP$_{\rm M}$ & AP$_{\rm L}$ \\

\midrule[1.0pt]

DETR-R50~\cite{DETR} $\ddag$&  & 50 & 41M & 86G & 24.6 & 2.1\,GB & 34.9 & 55.5 & 36.0 & 14.4 & 37.2 & 54.5 \\

DETR-R50-DC5~\cite{DETR} $\ddag$&  $\checkmark$ & 50 & 41M & 187G & 9.2 & 5.8\,GB & 36.7 & 57.6 & 38.2 & 15.4 & 39.8 & \textbf{56.3} \\

\rowcolor{Gray} DETR-R50~\cite{DETR} + IMFA\,(Ours) $\ddag$ & & 50 & 52M & 105G & 20.0 & 2.5\,GB & \textbf{39.2} & \textbf{58.8} & \textbf{41.6} & \textbf{20.3} & \textbf{42.2} & 55.4 \\

\midrule[0.6pt]

Conditional-DETR-R50~\cite{ConditionalDETR} & & 50 & 44M & 90G & 22.2 & 2.1\,GB & 40.9 & 61.8 & 43.3 & 20.8 & 44.6 & 59.2 \\

Conditional-DETR-R50-DC5~\cite{ConditionalDETR} & $\checkmark$ & 50 & 44M & 195G & 8.9 & 5.8\,GB & 43.8 & \textbf{64.4} & 46.7 & 24.0 & \textbf{47.6} & \textbf{60.7} \\

\rowcolor{Gray} Conditional-DETR-R50\,\cite{ConditionalDETR}\,+\,IMFA\,(Ours) & & 50 & 53M & 106G & 19.0 & 2.5\,GB & \textbf{44.0} & 64.2 & \textbf{47.5} & \textbf{25.7} & 46.8 & 59.8 \\

\midrule[0.6pt]

Anchor-DETR-R50~\cite{AnchorDETR} &  & 50 & 37M & 93G & 22.3 & 2.1\,GB & 42.1 & 63.1 & 44.9 & 22.3 & 46.2 & 60.0 \\

Anchor-DETR-R50-DC5~\cite{AnchorDETR} & $\checkmark$ & 50 & 37M & 172G & 14.3 & 3.6\,GB & 44.2 & \textbf{64.7} & 47.5 & 24.7 & \textbf{48.2} & \textbf{60.6} \\

\rowcolor{Gray} Anchor-DETR-R50~\cite{AnchorDETR} + IMFA\,(Ours) & & 50 & 46M & 106G & 17.5 & 2.4\,GB & \textbf{44.5} & 63.9 & \textbf{47.7} &  \textbf{26.4} & 47.7 & 59.9  \\

\midrule[0.6pt]

DAB-DETR-R50~\cite{DABDETR} & & 50 & 44M & 94G & 21.4 & 2.1\,GB & 42.2 & 63.1 & 44.7 & 21.5 &45.7 & 60.3 \\

DAB-DETR-R50-DC5~\cite{DABDETR} & $\checkmark$ & 50 & 44M & 202G & 8.8 & 6.0\,GB & 44.5 & \textbf{65.1} &47.7 &25.3 &48.2 & \textbf{62.3} \\

\rowcolor{Gray} DAB-DETR-R50~\cite{DABDETR} + IMFA\,(Ours) & & 50 & 53M & 108G & 18.6 & 2.5\,GB & \textbf{45.5} & 65.0 & \textbf{49.3} & \textbf{27.3} & \textbf{48.3} & 61.6 \\

\bottomrule[1.5pt]
\end{tabular}}
\vspace{-2mm}
\caption{Compatibility with different Transformer-based object detectors. IMFA boosts the performance of existing detectors at slight computational costs. `High-Res\,Feat' denotes the use of high-resolution features with R50-DC5. \,$\ddag$~denotes DETR with 300 object queries and focal loss. Results are reported on COCO val\,2017.
}
\label{tab:exp_easy_integration}
\vspace{-4.486mm}
\end{center}
\end{table*}

\vspace{+1.2mm}
\noindent
\textbf{Sampling Scale-Adaptive Features from Representative Keypoints.\;}
IMFA directly samples multi-scale features from the feature pyramid that is generated from the backbone (C2-C5 from ResNet in our experiments). However, even the sparsely sampled promising regions still contain a substantial amount of feature tokens at high-resolution feature scales. To further sparsify the sampled multi-scale features, IMFA searches a small number of representative keypoints within each promising region and samples their corresponding features at adaptively selected scales.

As illustrated in Fig.~\ref{fig:IMFA}, for each promising region, IMFA first uses its object query to predict $M$ keypoint locations within the region, which can be formulated as:
\begin{equation}
    \{P_{ij}\}_{j=1}^M = \text{MLP}(\mathbf{Q}_i) \;\;\;\; \text{for} \; i= 1, 2, ..., K\;,
\end{equation}
where $i$ and $j$ index the queries and keypoints, respectively, and each keypoint $P_{ij} = (x_{ij}, y_{ij})$ lies within its corresponding box prediction $\mathbf{B}_i$. Then, IMFA samples each keypoint's features from the feature pyramid at all scales via bilinear interpolation, obtaining a set of features $\{\mathbf{F}_{ij}^s\}_{s=1}^S$, where $S$ is the number of feature scales. Finally, to emphasize the distinct significance of different feature scales for each keypoint, we propose to perform adaptive scale selection by predicting scale-specific weights for each keypoint and obtaining scale-adaptive features through weighted summation:
\begin{equation}
   \mathbf{F}_{ij} = \textstyle \sum_{s} \alpha_{ij}^s \mathbf{F}_{ij}^s \;\;\;\;\;\;\; \{\alpha_{ij}^s\}_{s=1}^S = \text{Softmax}(\gamma_j(\mathbf{Q}_i)),
\end{equation}
where the scale-selection weights $\alpha$ are generated by a linear projection $\gamma_j$ followed by a Softmax function, so that $\sum_{s}\alpha_{ij}^{s}=1$. 
In this way, IMFA only samples the most crucial and informative features, producing a set of sparse yet still highly informative multi-scale features for each promising region. Additionally, to further strengthen the representation capacity of the sampled multi-scale features, we feed the sampled features into a Dynamic Feed-Forward Network (Dynamic FFN) to incorporate the semantics from their corresponding object queries via dynamic weighting~\cite{SparseRCNN}, where FFN's weights are dynamically generated by object queries. It can be formulated as:
\begin{equation}
    \mathbf{F}_{ij}' = \text{MLP}_{\mathbf{W}_{i}}(\mathbf{F}_{ij})   \;\;\;\;\;\;\;\;\; \mathbf{W}_{i} = \psi(\mathbf{Q}_i).
\end{equation}
Here, for each object query $\mathbf{Q}_i$, the dynamic weight $\mathbf{W}_i$ is obtained by a linear projection $\psi$ of $\mathbf{Q}_i$. Then, $\mathbf{W}_i$ is applied to the scale-adaptive features $\mathbf{F}_{ij}$ to generate the final sampled features $\mathbf{F}_{ij}'$ with enhanced semantics. These sampled features, along with their positional embeddings obtained based on their keypoint locations, are further fed to the subsequent detection stage for aggregation.

\vspace{+1.2mm}
\noindent
\textbf{Iterative Aggregation of Multi-Scale Features.\;}
To leverage the sampled multi-scale features for refined object detection, the sampled features and the encoded image features are fed into the subsequent encoder layer for aggregation using the attention mechanism. This is analogous to the top-down path created by FPN~\cite{FPN} for enhancing the semantics of low-level features. To avoid continuous growth of feature tokens and maintain efficiency, each detection stage does not inherit the multi-scale features that are generated from the previous stage, as shown in Fig.~\ref{fig:IMFA}.

\subsection{Visualization and Analysis}      \label{sec:IMFA_method_vis_analysis}

Fig.~\ref{fig:IMFA_vis} visualizes IMFA's sampling locations and their feature scales. It can be observed that the sampling locations mostly fall around the target objects, and typically at representative locations, such as object extremities. This proves the effectiveness of IMFA in searching sparse yet highly informative locations in the feature sampling process. Besides, it is noteworthy that IMFA tends to focus on higher-resolution features for small objects and lower-resolution features for large objects, which is intuitive as the detection of small objects relies more on finer details.

\section{Experiments}           \label{sec:IMFA_exp}

\subsection{Experiment Setup}        \label{sec:IMFA_exp_setups}

\noindent
\textbf{Dataset and Evaluation Metrics.\;}
We perform experiments on the COCO\,2017 dataset~\cite{MSCOCO}. We use $\sim$117k images in \verb+train2017+ for training and 5k images in \verb+val2017+ for evaluation. We adopt COCO's standard evaluation metrics for performance evaluation.

\vspace{+1.0mm}
\noindent
\textbf{Implementation Details.\;}
As the proposed IMFA defines a generic paradigm, we mainly conduct experiments with DAB-DETR~\cite{DABDETR} -- a state-of-the-art Transformer-based object detector with open-sourced implementation. We also integrate IMFA with DETR~\cite{DETR}, Conditional DETR~\cite{ConditionalDETR}, and Anchor DETR~\cite{AnchorDETR}, to demonstrate its generality.

A crucial implementation detail involves incorporating skip connections for encoded features between Transformer encoder layers, as motivated by \cite{Meta-DETR-firstversion} and ~\cite{SAM-DETR,SAM-DETR-PlusPlus} to facilitate feature semantic alignment.

For IMFA-related hyper-parameters, we set the sampling ratio $r$ at 20\% and the keypoint number $M$ at 8 by default. Other model-related setups align with their corresponding baselines~\cite{DETR,ConditionalDETR,AnchorDETR,DABDETR}. We use ImageNet-pretrained\cite{imagenet} ResNet\cite{resnet} as backbone networks, and conduct training with AdamW optimizer~\cite{AdamW}. The total batch size is set to 16 for training. The initial learning rate is $\rm 1\!\times\!10^{-5}$ for the backbone networks and $\rm 1\!\times\!10^{-4}$ for the Transformer architectures, along with a weight decay of $\rm 1\!\times\!10^{-4}$. Models are trained for 50 epochs, with the learning rate decayed at the 40$^{\rm th}$ epoch by 0.1. The same data augmentation scheme used in \cite{DETR,ConditionalDETR,AnchorDETR,DABDETR} is adopted.

\begin{table*}[t]
\begin{center}
\centering
\vspace{-2.0mm}
\setlength{\tabcolsep}{6.486pt}
\resizebox{0.92\textwidth}{!}{
\begin{tabular}[t]{l|ccc|ccc|cccccc}

\toprule[1.5pt]
Method & MS & SMS & DC & \#Epochs & \#Params & FLOPs & AP & AP$_{\rm 50}$ & AP$_{\rm 75}$ & AP$_{\rm S}$ & AP$_{\rm M}$ & AP$_{\rm L}$ \\

\midrule[1.0pt]

Faster-RCNN-FPN-R50~\cite{FasterRCNN,FPN} & $\checkmark$ & &  & 108 & 42M & 180G & 42.0 & 62.1 & 45.5 & 26.6 & 45.5 & 53.4  \\

TSP-FCOS-FPN-R50~\cite{TSPRCNN} & $\checkmark$ & & & 36 & 52M & 189G & 43.1 & 62.3 & 47.0 & 26.6 & 46.8 & 55.9 \\

TSP-RCNN-FPN-R50~\cite{TSPRCNN} & $\checkmark$ & & & 36 & 64M & 188G & 43.8 & 63.3 & 48.3 & \textbf{28.6} & 46.9 & 55.7 \\

Sparse-RCNN-FPN-R50~\cite{SparseRCNN} & $\checkmark$ & & & 36 & 106M & 166G & 45.0 & 64.1 & 48.9 & 28.0 & 47.6 & 59.5 \\

DETR-R50~\cite{DETR} & & & $\checkmark$ & 500 & 41M & 187G & 43.3 & 63.1 & 45.9 & 22.5 & 47.3 & 61.1 \\

Deformable-DETR-R50~\cite{DeformableDETR} & $\checkmark$ & &  & 50 & 40M & 173G & 43.8 & 62.6 & 47.7 & 26.4 & 47.1 & 58.0 \\

Deformable-DETR-R50~\cite{DeformableDETR}\,+\,Iter & $\checkmark$ & &  & 50 & 41M & 173G & 45.4 & 64.7 & 49.0 & 26.8 & 48.3 & 61.7 \\

Efficient-DETR-R50~\cite{EfficientDETR} & $\checkmark$ & & & 36 & 32M & 159G & 44.2 & 62.2 &48.0 & 28.4 &47.5 & 56.6 \\

Conditional-DETR-R50~\cite{ConditionalDETR} & & & $\checkmark$ & 50 & 44M & 195G & 43.8 & 64.4 & 46.7 & 24.0 & 47.6 & 60.7 \\

SMCA-DETR-R50~\cite{SMCA-DETR} & $\checkmark$ & & & 50 & 40M & 152G & 43.7 & 63.6 & 47.2 & 24.2 & 47.0 & 60.4 \\

YOLOS-DeiT-S~\cite{YOLOS} &  &  &  & 150 & 28M & 172G & 37.6 & 57.6 & 39.2 & 15.9 & 40.2 & 57.3 \\

Anchor-DETR-R50~\cite{AnchorDETR} & & & $\checkmark$ & 50 & 37M & 172G & 44.2 & 64.7 & 47.5 & 24.7 & 48.2 & 60.6 \\

DAB-DETR-R50~\cite{DABDETR} & & & $\checkmark$ & 50 & 44M & 202G & 44.5 & 65.1 &47.7 &25.3 &48.2 & 62.3 \\

SAM-DETR-R50~\cite{SAM-DETR} & & & $\checkmark$ & 50 & 58M & 210G & 43.3 & 64.4 & 46.2 & 25.1 & 46.9 & 61.0 \\

SAM-DETR-R50\,\cite{SAM-DETR}\;w/\;SMCA~\cite{SMCA-DETR} & & & $\checkmark$ & 50 & 58M & 210G & 45.0 & \textbf{65.4} & 47.9 & 26.2 & \textbf{49.0} & \textbf{63.3}  \\

\rowcolor{Gray} DAB-DETR-R50~\cite{DABDETR}\,+\,IMFA\,(Ours) & & $\checkmark$ &  & 50 & 53M & \textbf{108G} & \textbf{45.5} & 65.0 & \textbf{49.3} & 27.3 & 48.3 & 61.6 \\

\bottomrule[1.5pt]
\end{tabular}}
\end{center}
\vspace{-5.5mm}
\caption{Comparison with state-of-the-art object detectors on COCO val\,2017. 
Our proposed method achieves comparable performance with the state-of-the-art methods, but with significantly lower computation.
`MS' denotes the use of multi-scale features. `SMS' denotes the use of sparse multi-scale features with our proposed IMFA. `DC' denotes the use of high-resolution features with R50-DC5.
}
\label{tab:exp_comparison_with_sota}
\vspace{-1.0mm}
\end{table*}

\begin{table}[t!]
\begin{center}
\centering
\setlength{\tabcolsep}{2.0pt}
\resizebox{0.4486\textwidth}{!}{
\begin{tabular}[t]{l|ccc|c}

\toprule[1.68pt]
Method & \#Params & FLOPs & FPS & AP \\

\midrule[1.1pt]

DETR-SwinB\cite{DETR} & 105M & 303G & 9.8 & 40.7 \\

\rowcolor{Gray} DETR-SwinB\cite{DETR}\,+\,IMFA\,(Ours) & 115M & 318G & 9.3 & \textbf{46.2} \\

\midrule[0.3pt]

DAB-DETR-ConvNextB\cite{DABDETR} & 108M & 287G & 9.4 & 47.4 \\

\rowcolor{Gray} DAB-DETR-ConvNextB\cite{DABDETR}\,+\,IMFA\,(Ours) & 117M & 301G & 8.7 & \textbf{50.0} \\

\bottomrule[1.68pt]
\end{tabular}}
\end{center}
\vspace{-5.6mm}
\caption{Results under stronger backbones. Results are obtained on COCO val\,2017.}
\label{tab:exp_stronger_backbones}
\vspace{-4mm}
\end{table}

\subsection{Experiment Results}       \label{sec:IMFA_exp_results}

\vspace{-1.0mm}
\noindent
\textbf{Compatibility with Transformer-Based Detectors.\;}
We first evaluate the generality of IMFA by integrating it with multiple Transformer-based object detectors.
As discussed in Section~\ref{sec:IMFA_introduction}, these methods resort to higher-resolution backbones (denoted with `High-Res Feat') as an alternative, as it is computationally prohibitive for them to directly process multi-scale features.
As shown in Table~\ref{tab:exp_easy_integration}, using higher-resolution features improves the detection performance but adds a substantial computational cost (+\,$\sim$100\,GFLOPs and -\,8$\sim$15\,FPS) as well as GPU memory consumption. On the other hand, the proposed IMFA consistently improves the detection performance by large margins across all metrics, especially on small objects ($\rm AP_{\rm S}$), yet only introduces a slight computational overhead (+\,$\rm \sim$15\,GFLOPs and -\,$\sim$3\,FPS). The experimental results demonstrate IMFA's effectiveness and wide applicability.

\vspace{+0.8mm}
\noindent
\textbf{Comparison with State-of-the-Art Detectors.\;}
We integrate IMFA with DAB-DETR~\cite{DABDETR} to benchmark with other state-of-the-art single-stage Transformer-based detectors that utilize high-resolution or multi-scale features. We also include some popular two-stage detectors~\cite{FasterRCNN,TSPRCNN,EfficientDETR} for a comprehensive comparison. As shown in Table~\ref{tab:exp_comparison_with_sota}, our method can achieve comparable performance with the state-of-the-art methods, but with significantly less computational cost.

\vspace{+0.8mm}
\noindent
\textbf{Results with Stronger Backbones.\;}
As shown in Table~\ref{tab:exp_stronger_backbones}, when using stronger backbones~\cite{SwinTransformer,convnext}, IMFA still consistently improves detection performance at marginal costs.

\begin{table}[t]
\begin{center}
\centering
\setlength{\tabcolsep}{4.8866pt}
\resizebox{0.4141\textwidth}{!}{
\begin{tabular}[t]{cc|cc|cccc}

\toprule[1.68pt]
Iter.\,Enc. & SFSA & \#Params & FLOPs & AP & AP$_{\rm S}$ & AP$_{\rm M}$ & AP$_{\rm L}$ \\

\midrule[1.1pt]

 & & 44M & 94G & 42.2 & 21.5 & 45.7 & 60.3 \\

$\checkmark$ & & 44M & 94G & 41.9 & 21.8 & 45.2 & 61.1 \\

\rowcolor{Gray} $\checkmark$ & $\checkmark$ & 53M & 108G & \textbf{45.5} & \textbf{27.3} & \textbf{48.3} & \textbf{61.6} \\

\bottomrule[1.68pt]
\end{tabular}}
\end{center}
\vspace{-5.60mm}
\caption{Ablation studies on IMFA's two major design choices. `Iter.\,Enc.' denotes iterative update of encoded features as illustrated in Fig.\,\ref{fig:difference}\,(right). `SFSA' denotes sparse feature sampling and aggregation as illustrated in Fig.\,\ref{fig:IMFA}.}
\label{tab:ablation_designs_1}
\vspace{-0.8mm}
\end{table}

\begin{table}[t]
\begin{center}
\centering
\setlength{\tabcolsep}{3pt}
\resizebox{0.4486\textwidth}{!}{
\begin{tabular}[t]{ccc|cc|cccc}

\toprule[1.68pt]
Rep.\,Kp. & Ada.\,Scale & Dy.\,FFN & \#Params & FLOPs & AP & AP$_{\rm S}$ & AP$_{\rm M}$ & AP$_{\rm L}$ \\

\midrule[1.1pt]

& & & 44M & 94G & 41.9 & 21.8 & 45.2 & 61.1 \\

& $\checkmark$ & & 45M & 105G & 42.1 & 22.0 & 45.4 & 61.0 \\

& $\checkmark$ & $\checkmark$ & 53M & 108G & 42.3 & 22.2 & 46.0 & 60.9 \\

$\checkmark$ & & $\checkmark$ & 53M & 108G & 44.7 & 26.4 & 47.6 & 61.5 \\

$\checkmark$ & $\checkmark$ & & 45M & 105G & 44.2 & 26.3 & 47.2 & 60.8 \\

\rowcolor{Gray} $\checkmark$ & $\checkmark$ & $\checkmark$ & 53M & 108G & \textbf{45.5} & \textbf{27.3} & \textbf{48.3} & \textbf{61.6} \\

\bottomrule[1.68pt]
\end{tabular}}
\end{center}
\vspace{-5.60mm}
\caption{Ablation studies on the design choices within sparse multi-scale feature sampling and aggregation. `Rep.\,Kp.' denotes searching representative keypoints. `Ada.\,Scale' denotes adaptive scale selection. `Dy.\,FFN' denotes Dynamic FFN.}
\label{tab:ablation_designs_2}
\vspace{-1.25mm}
\end{table}

\subsection{Ablation Study}       \label{sec:IMFA_exp_ablation}

\vspace{-0.66mm}
We conduct ablation studies with the strong baseline DAB-DETR-R50~\cite{DABDETR,resnet} to validate the effectiveness of our designs. Results are obtained on COCO val\,2017.

\vspace{+1mm}
\noindent
\textbf{Effect of IMFA's Design Choices.\;}
IMFA introduces two novel designs: \textit{i)} iterative encoding described in Section~\ref{sec:IMFA_method_iter_update} and Fig.~\ref{fig:difference}\,(right), and \textit{ii)} sparse multi-scale feature sampling and aggregation described in Section~\ref{sec:IMFA_method_sparse} and Fig.~\ref{fig:IMFA}. As shown in Table~\ref{tab:ablation_designs_1}, the iterative encoding alone even slightly degrades the baseline's performance. However, with IMFA's sparsely sampled multi-scale features, our method significantly improves the detection performance of objects at all scales, especially at smaller scales. This proves that the multi-scale features sampled by IMFA are sparse yet highly effective for object detection.

We also study the three crucial components within the sparse feature sampling and aggregation process in Table~\ref{tab:ablation_designs_2}.
Without identifying representative keypoints (random spatial sampling is used instead), the performance barely improves, which verifies our claim that only a very small set of multi-scale features are beneficial. The results also validate that IMFA can search keypoints with important semantics information. Without adaptive scale selection (averaged scale selection is used instead), the performance drops, indicating that our design enables the focus of appropriate scales for each object. Without Dynamic FFN, the performance also drops, which proves that Dynamic FFN successfully fuses important semantics information from the corresponding object queries and benefits the final prediction.

\vspace{+1.0mm}
\noindent
\textbf{Effect of IMFA's Hyper-Parameters.\;}
IMFA introduces two hyper-parameters: the sampling ratio of prior detection predictions and object queries ($r$) as well as the keypoint number in each promising region ($M$). We conduct sensitivity analysis on each of them.


\begin{table}[t!]
\begin{center}
\centering
\setlength{\tabcolsep}{4.86pt}
\resizebox{0.91\linewidth}{!}{
\begin{tabular}[t]{c|cc|cccccc}

\toprule[1.6pt]
$r$ & \#Params & FLOPs & AP & AP$_{\rm 50}$ & AP$_{\rm 75}$ & AP$_{\rm S}$ & AP$_{\rm M}$ & AP$_{\rm L}$ \\

\midrule[1.0pt]

10\% & 53M & 103G &  44.2 & 64.0 & 47.5 & 25.9 & 47.3 & 60.6 \\

15\% & 53M & 105G & 44.8 & 64.2 & 48.2 & 26.5 & 47.7 & 60.1 \\

\rowcolor{Gray} 20\% & 53M & 108G & \textbf{45.5} & 65.0 & \textbf{49.3} & 27.3 & \textbf{48.3} & \textbf{61.6} \\

25\% & 53M & 111G & 45.3 & \textbf{65.1} & 49.0 & 27.9 & 47.9 & 61.1 \\

30\% & 53M & 114G & 45.1 & 64.5 & 48.9 & \textbf{28.4} & 48.2 & 60.2 \\

\bottomrule[1.6pt]
\end{tabular}}
\vspace{-2mm}
\caption{Ablation study on the sampling ratio $r$ of prior detection predictions. Results are obtained on COCO val\,2017.}
\label{tab:ablation_percentage}
\vspace{-4.86mm}
\end{center}
\end{table}

\begin{table}[t]
\centering
\begin{center}
\centering
\setlength{\tabcolsep}{4.86pt}
\resizebox{0.91\linewidth}{!}{
\begin{tabular}[t]{c|cc|cccccc}

\toprule[1.6pt]
$M$ & \#Params & FLOPs & AP & AP$_{\rm 50}$ & AP$_{\rm 75}$ & AP$_{\rm S}$ & AP$_{\rm M}$ & AP$_{\rm L}$ \\

\midrule[1.0pt]

1 & 53M & 101G &  43.9 & 64.3 & 47.5 & 25.1 & 46.9 & 60.8 \\

2 & 53M & 102G & 45.0 & 64.7 & 48.9 & 26.0 & 48.3 & 60.4 \\

4 & 53M & 104G & 45.3 & \textbf{65.0} & 48.7 & \textbf{27.3} & 48.1 & 60.9 \\

\rowcolor{Gray} 8 & 53M & 108G & \textbf{45.5} & \textbf{65.0} & \textbf{49.3} & \textbf{27.3} & 48.3 & \textbf{61.6}  \\

16 & 53M & 117G &  45.3 & 64.7 & 49.0 & 26.6 & \textbf{48.5} & 61.5  \\

\bottomrule[1.6pt]
\end{tabular}}
\vspace{-2mm}
\caption{Ablation study on the keypoint number $M$ within each promising region. Results are obtained on COCO val\,2017.}
\label{tab:ablation_num_keypoint}
\vspace{-4.86mm}
\end{center}
\end{table}


Table~\ref{tab:ablation_percentage} shows the effect of different $r$ values when $M$ is fixed at 8.
As $r$ increases from 10\% to 30\%, the average precision (AP) first increases then decreases, while the computational cost keeps growing. An interesting trend is that the detection performance of small objects (AP\textsubscript{S}) goes up with increasing $r$ consistently. We conjecture that small objects rely more on the fine details in high-resolution features, so that they can benefit from increased number of promising regions used for multi-scale feature sampling. However, the overall performance drops when $r$ is too large, which we conjecture is due to the increased difficulty in searching relevant features with overwhelming featuare tokens involved. Based on the experimental results, we set the default value for $r$ as 20\% in our system.

To study the effect of the number of keypoints $M$, we conduct experiments by fixing $r$ at 20\% and report the results in Table~\ref{tab:ablation_num_keypoint}. We can see a similar trend that the performance improves as $M$ increases but then drops when $M$ becomes too large. Therefore, we set $M$ as 8 by default.

\subsection{Extension to Human Pose Estimation}       \label{sec:IMFA_exp_kpt_det}

We further apply the proposed IMFA to human pose estimation to verify its generality across different tasks. Concretely, we evaluate the performance on the COCO\,2017 human pose estimation benchmark~\cite{MSCOCO}. We adopt PRTR (two-stage variant)\cite{PRTR}, a DETR-like human pose estimation method with open-sourced implementation, as our baseline. Please refer to the technical appendix for its full implementation details.

\begin{table}[t!]
\begin{center}
\centering
\setlength{\tabcolsep}{3.9999pt}
\resizebox{0.4486\textwidth}{!}{
\begin{tabular}[t]{l|c|cc|c}

\toprule[1.38pt]
Method & Input\,Size & FLOPs & FPS &  AP$^{\rm kp}$  \\

\midrule[1.0pt]

PRTR-R50\cite{PRTR} & 384x288 & 11.0\,G & 360 & 68.2 \\

PRTR-R50\cite{PRTR} & 512x384 & 18.8\,G & 218 & 71.0 \\

\rowcolor{Gray} PRTR-R50\cite{PRTR}\,+\,IMFA\,(Ours) & 384x288 & 13.4\,G & 293 & \textbf{72.7} \\

\midrule[0.2pt]

PRTR-R101\cite{PRTR} & 384x288 & 19.1\,G & 243 & 70.1 \\

PRTR-R101\cite{PRTR} & 512x384 & 33.4\,G & 144 & 72.0 \\

\rowcolor{Gray} PRTR-R101\cite{PRTR}\,+\,IMFA\,(Ours) & 384x288 & 21.5\,G & 216 & \textbf{73.7} \\

\bottomrule[1.38pt]
\end{tabular}}
\end{center}
\vspace{-5.25mm}
\caption{Human pose estimation performance on COCO val\,2017. IMFA greatly boosts performance at marginal costs, even surpassing the baseline methods with high-resolution input images.}
\label{tab:exp_keypoint_detection}
\vspace{-1.0mm}
\end{table}

As shown in Table~\ref{tab:exp_keypoint_detection}, on the task of human pose estimation, IMFA still clearly outperforms its baseline methods at the same input size with only slight extra computation. IMFA even surpasses its higher-resolution baselines at significantly reduced computational costs. The results indicate IMFA's potential of boosting Transformer-based models on various vision tasks beyond object detection itself.

\section{Conclusion}

Multi-scale features are beneficial to object detection, but often come with large computational costs. This paper presents \textit{Iterative Multi-scale Feature Aggregation (IMFA)} as the pioneering generic paradigm for efficient use of multi-scale features in Transformer-based object detectors. It gets the best of both worlds -- high accuracy and low computational cost. IMFA identifies and extracts multi-scale features from the most promising and informative locations only and greatly improves detection accuracy on multiple object detectors at marginal additional costs. We expect IMFA will inspire more comprehensive research and applications on Transformer-based object detection.

\vspace{+1.125mm}
\noindent
\textbf{Limitations.\;}
Although IMFA is compatible with many Transformer-based object detectors, it cannot be directly applied to Deformable DETR~\cite{DeformableDETR} and its extensions~\cite{SparseDETR,EfficientDETR}. This is due to undefined deformable operations on non-grid feature maps, which require extensive engineering efforts.

\vspace{+1.25mm}
\section*{\normalsize Acknowledgement:}
\vspace{-2.5mm}
This study is supported under the RIE\,2020 Industry Alignment Fund – Industry Collaboration Projects (IAF-ICP) Funding Initiative, as well as cash and in-kind contribution from the industry partner(s).

\clearpage
{\small
\bibliographystyle{ieee_fullname}
\bibliography{egbib}

\begin{thebibliography}{10}\itemsep=-1pt

\bibitem{detreg}
Amir Bar, Xin Wang, Vadim Kantorov, Colorado~J Reed, Roei Herzig, Gal Chechik,
  Anna Rohrbach, Trevor Darrell, and Amir Globerson.
\newblock {DETReg}: Unsupervised pretraining with region priors for object
  detection.
\newblock In {\em CVPR}, 2022.

\bibitem{CascadeRCNN}
Zhaowei Cai and Nuno Vasconcelos.
\newblock {Cascade R-CNN}: Delving into high quality object detection.
\newblock In {\em CVPR}, 2018.

\bibitem{CF-DETR}
Xipeng Cao, Peng Yuan, Bailan Feng, and Kun Niu.
\newblock {CF-DETR}: Coarse-to-fine transformers for end-to-end object
  detection.
\newblock In {\em AAAI}, 2022.

\bibitem{DETR}
Nicolas Carion, Francisco Massa, Gabriel Synnaeve, Nicolas Usunier, Alexander
  Kirillov, and Sergey Zagoruyko.
\newblock End-to-end object detection with {T}ransformers.
\newblock In {\em ECCV}, 2020.

\bibitem{up-detr}
Zhigang Dai, Bolun Cai, Yugeng Lin, and Junying Chen.
\newblock {UP-DETR}: Unsupervised pre-training for object detection with
  transformers.
\newblock In {\em CVPR}, 2021.

\bibitem{imagenet}
Jia Deng, Wei Dong, Richard Socher, Li-Jia Li, Kai Li, and Li Fei-Fei.
\newblock {ImageNet}: A large-scale hierarchical image database.
\newblock In {\em CVPR}, 2009.

\bibitem{fsod}
Qi Fan, Wei Zhuo, Chi-Keung Tang, and Yu-Wing Tai.
\newblock Few-shot object detection with attention-{RPN} and multi-relation
  detector.
\newblock In {\em CVPR}, 2020.

\bibitem{YOLOS}
Yuxin Fang, Bencheng Liao, Xinggang Wang, Jiemin Fang, Jiyang Qi, Rui Wu,
  Jianwei Niu, and Wenyu Liu.
\newblock You only look at one sequence: Rethinking transformer in vision
  through object detection.
\newblock In {\em NeurIPS}, 2021.

\bibitem{figurnov2017spatially}
Michael Figurnov, Maxwell~D Collins, Yukun Zhu, Li Zhang, Jonathan Huang,
  Dmitry Vetrov, and Ruslan Salakhutdinov.
\newblock Spatially adaptive computation time for residual networks.
\newblock In {\em CVPR}, 2017.

\bibitem{perforatedCNNs}
Mikhail Figurnov, Aizhan Ibraimova, Dmitry~P Vetrov, and Pushmeet Kohli.
\newblock {PerforatedCNNs}: Acceleration through elimination of redundant
  convolutions.
\newblock In {\em NeurIPS}, 2016.

\bibitem{SMCA-DETR}
Peng Gao, Minghang Zheng, Xiaogang Wang, Jifeng Dai, and Hongsheng Li.
\newblock Fast convergence of {DETR} with spatially modulated co-attention.
\newblock In {\em ICCV}, 2021.

\bibitem{NASFPN}
Golnaz Ghiasi, Tsung-Yi Lin, and Quoc~V Le.
\newblock {NAS-FPN}: Learning scalable feature pyramid architecture for object
  detection.
\newblock In {\em CVPR}, 2019.

\bibitem{OW-DETR}
Akshita Gupta, Sanath Narayan, KJ Joseph, Salman Khan, Fahad~Shahbaz Khan, and
  Mubarak Shah.
\newblock {OW-DETR}: Open-world detection transformer.
\newblock In {\em CVPR}, 2022.

\bibitem{MaskRCNN}
Kaiming He, Georgia Gkioxari, Piotr Doll{\'a}r, and Ross Girshick.
\newblock {Mask R-CNN}.
\newblock In {\em ICCV}, 2017.

\bibitem{resnet}
Kaiming He, Xiangyu Zhang, Shaoqing Ren, and Jian Sun.
\newblock Deep residual learning for image recognition.
\newblock In {\em CVPR}, 2016.

\bibitem{MDETR}
Aishwarya Kamath, Mannat Singh, Yann LeCun, Gabriel Synnaeve, Ishan Misra, and
  Nicolas Carion.
\newblock {MDETR}--modulated detection for end-to-end multi-modal
  understanding.
\newblock In {\em ICCV}, 2021.

\bibitem{FewshotReweighting}
Bingyi Kang, Zhuang Liu, Xin Wang, Fisher Yu, Jiashi Feng, and Trevor Darrell.
\newblock Few-shot object detection via feature reweighting.
\newblock In {\em ICCV}, 2019.

\bibitem{parallelfpn}
Seung-Wook Kim, Hyong-Keun Kook, Jee-Young Sun, Mun-Cheon Kang, and Sung-Jea
  Ko.
\newblock Parallel feature pyramid network for object detection.
\newblock In {\em ECCV}, 2018.

\bibitem{PanopticFPN}
Alexander Kirillov, Ross Girshick, Kaiming He, and Piotr Doll{\'a}r.
\newblock Panoptic feature pyramid networks.
\newblock In {\em CVPR}, 2019.

\bibitem{DN-DETR}
Feng Li, Hao Zhang, Shilong Liu, Jian Guo, Lionel~M Ni, and Lei Zhang.
\newblock {DN-DETR}: Accelerate {DETR} training by introducing query denoising.
\newblock In {\em CVPR}, 2022.

\bibitem{PRTR}
Ke Li, Shijie Wang, Xiang Zhang, Yifan Xu, Weijian Xu, and Zhuowen Tu.
\newblock Pose recognition with cascade transformers.
\newblock In {\em CVPR}, 2021.

\bibitem{tridentnet}
Yanghao Li, Yuntao Chen, Naiyan Wang, and Zhaoxiang Zhang.
\newblock Scale-aware trident networks for object detection.
\newblock In {\em ICCV}, 2019.

\bibitem{MapTR}
Bencheng Liao, Shaoyu Chen, Xinggang Wang, Tianheng Cheng, Qian Zhang, Wenyu
  Liu, and Chang Huang.
\newblock {MapTR}: Structured modeling and learning for online vectorized {HD}
  map construction.
\newblock In {\em ICLR}, 2023.

\bibitem{masktextspotter}
Minghui Liao, Pengyuan Lyu, Minghang He, Cong Yao, Wenhao Wu, and Xiang Bai.
\newblock {Mask TextSpotter}: An end-to-end trainable neural network for
  spotting text with arbitrary shapes.
\newblock {\em IEEE Transactions on Pattern Analysis and Machine Intelligence},
  43(2):532--548, 2021.

\bibitem{FPN}
Tsung-Yi Lin, Piotr Doll{\'a}r, Ross Girshick, Kaiming He, Bharath Hariharan,
  and Serge Belongie.
\newblock Feature pyramid networks for object detection.
\newblock In {\em CVPR}, 2017.

\bibitem{focalloss}
Tsung-Yi Lin, Priya Goyal, Ross Girshick, Kaiming He, and Piotr Doll{\'a}r.
\newblock Focal loss for dense object detection.
\newblock In {\em ICCV}, 2017.

\bibitem{MSCOCO}
Tsung-Yi Lin, Michael Maire, Serge~J. Belongie, Lubomir~D. Bourdev, Ross~B.
  Girshick, James Hays, Pietro Perona, Deva Ramanan, Piotr Doll{\'a}r, and
  C.~Lawrence Zitnick.
\newblock Microsoft {COCO}: Common objects in context.
\newblock In {\em ECCV}, 2014.

\bibitem{Liu2019DeepLF}
Li Liu, Wanli Ouyang, Xiaogang Wang, Paul Fieguth, Jie Chen, Xinwang Liu, and
  Matti Pietik{\"a}inen.
\newblock Deep learning for generic object detection: A survey.
\newblock {\em International Journal of Computer Vision}, 128:261--318, 2020.

\bibitem{DABDETR}
Shilong Liu, Feng Li, Hao Zhang, Xiao Yang, Xianbiao Qi, Hang Su, Jun Zhu, and
  Lei Zhang.
\newblock {DAB}-{DETR}: Dynamic anchor boxes are better queries for {DETR}.
\newblock In {\em ICLR}, 2022.

\bibitem{PANet}
Shu Liu, Lu Qi, Haifang Qin, Jianping Shi, and Jiaya Jia.
\newblock Path aggregation network for instance segmentation.
\newblock In {\em CVPR}, 2018.

\bibitem{SwinTransformer}
Ze Liu, Yutong Lin, Yue Cao, Han Hu, Yixuan Wei, Zheng Zhang, Stephen Lin, and
  Baining Guo.
\newblock {Swin Transformer}: Hierarchical vision {Transformer} using shifted
  windows.
\newblock In {\em ICCV}, 2021.

\bibitem{convnext}
Zhuang Liu, Hanzi Mao, Chao-Yuan Wu, Christoph Feichtenhofer, Trevor Darrell,
  and Saining Xie.
\newblock A {ConvNet} for the 2020s.
\newblock In {\em CVPR}, 2022.

\bibitem{AdamW}
Ilya Loshchilov and Frank Hutter.
\newblock Decoupled weight decay regularization.
\newblock In {\em ICLR}, 2019.

\bibitem{TransPillars}
Zhipeng Luo, Gongjie Zhang, Changqing Zhou, Tianrui Liu, Shijian Lu, and Liang
  Pan.
\newblock {TransPillars}: Coarse-to-fine aggregation for multi-frame {3D}
  object detection.
\newblock In {\em WACV}, 2023.

\bibitem{ConditionalDETR}
Depu Meng, Xiaokang Chen, Zejia Fan, Gang Zeng, Houqiang Li, Yuhui Yuan, Lei
  Sun, and Jingdong Wang.
\newblock Conditional {DETR} for fast training convergence.
\newblock In {\em ICCV}, 2021.

\bibitem{3DETR}
Ishan Misra, Rohit Girdhar, and Armand Joulin.
\newblock {An End-to-End Transformer Model for {3D} Object Detection}.
\newblock In {\em ICCV}, 2021.

\bibitem{autofocus}
Mahyar Najibi, Bharat Singh, and Larry~S Davis.
\newblock {AutoFocus}: Efficient multi-scale inference.
\newblock In {\em ICCV}, 2019.

\bibitem{LibraRCNN}
Jiangmiao Pang, Kai Chen, Jianping Shi, Huajun Feng, Wanli Ouyang, and Dahua
  Lin.
\newblock Libra {R-CNN}: Towards balanced learning for object detection.
\newblock In {\em CVPR}, 2019.

\bibitem{qiu2020borderdet}
Han Qiu, Yuchen Ma, Zeming Li, Songtao Liu, and Jian Sun.
\newblock {BorderDet}: Border feature for dense object detection.
\newblock In {\em ECCV}, 2020.

\bibitem{YOLO9000}
Joseph Redmon and Ali Farhadi.
\newblock {YOLO 9000}: Better, faster, stronger.
\newblock In {\em CVPR}, 2017.

\bibitem{SBNet}
Mengye Ren, Andrei Pokrovsky, Bin Yang, and Raquel Urtasun.
\newblock {SBNet}: Sparse blocks network for fast inference.
\newblock In {\em CVPR}, 2018.

\bibitem{FasterRCNN}
Shaoqing Ren, Kaiming He, Ross Girshick, and Jian Sun.
\newblock Faster {R-CNN}: Towards real-time object detection with region
  proposal networks.
\newblock In {\em NeurIPS}, 2015.

\bibitem{SparseDETR}
Byungseok Roh, JaeWoong Shin, Wuhyun Shin, and Saehoon Kim.
\newblock Sparse {DETR}: Efficient end-to-end object detection with learnable
  sparsity.
\newblock In {\em ICLR}, 2022.

\bibitem{StrongWeak}
Kuniaki Saito, Yoshitaka Ushiku, Tatsuya Harada, and Kate Saenko.
\newblock Strong-weak distribution alignment for adaptive object detection.
\newblock In {\em CVPR}, 2019.

\bibitem{ViDT}
Hwanjun Song, Deqing Sun, Sanghyuk Chun, Varun Jampani, Dongyoon Han, Byeongho
  Heo, Wonjae Kim, and Ming-Hsuan Yang.
\newblock {ViDT}: An efficient and effective fully transformer-based object
  detector.
\newblock In {\em ICLR}, 2022.

\bibitem{SparseRCNN}
Peize Sun, Rufeng Zhang, Yi Jiang, Tao Kong, Chenfeng Xu, Wei Zhan, Masayoshi
  Tomizuka, Lei Li, Zehuan Yuan, Changhu Wang, and Ping Luo.
\newblock {Sparse R-CNN}: End-to-end object detection with learnable proposals.
\newblock In {\em CVPR}, 2021.

\bibitem{TSPRCNN}
Zhiqing Sun, Shengcao Cao, Yiming Yang, and Kris~M. Kitani.
\newblock Rethinking {Transformer}-based set prediction for object detection.
\newblock In {\em ICCV}, 2021.

\bibitem{efficientdet}
Mingxing Tan, Ruoming Pang, and Quoc~V Le.
\newblock {EfficientDet}: Scalable and efficient object detection.
\newblock In {\em CVPR}, 2020.

\bibitem{FCOS}
Zhi Tian, Chunhua Shen, Hao Chen, and Tong He.
\newblock {FCOS}: Fully convolutional one-stage object detection.
\newblock In {\em ICCV}, 2019.

\bibitem{transformer}
Ashish Vaswani, Noam Shazeer, Niki Parmar, Jakob Uszkoreit, Llion Jones,
  Aidan~N. Gomez, L. Kaiser, and Illia Polosukhin.
\newblock Attention is all you need.
\newblock In {\em NeurIPS}, 2017.

\bibitem{verelst2020dynamic}
Thomas Verelst and Tinne Tuytelaars.
\newblock Dynamic convolutions: Exploiting spatial sparsity for faster
  inference.
\newblock In {\em CVPR}, 2020.

\bibitem{PnPDETR}
Tao Wang, Li Yuan, Yunpeng Chen, Jiashi Feng, and Shuicheng Yan.
\newblock {PnP-DETR}: Towards efficient visual analysis with {Transformers}.
\newblock In {\em ICCV}, 2021.

\bibitem{FP-DETR}
Wen Wang, Yang Cao, Jing Zhang, and Dacheng Tao.
\newblock {FP-DETR}: Detection transformer advanced by fully pre-training.
\newblock In {\em ICLR}, 2022.

\bibitem{fsdet}
Xin Wang, Thomas~E. Huang, Trevor Darrell, Joseph~E Gonzalez, and Fisher Yu.
\newblock Frustratingly simple few-shot object detection.
\newblock In {\em ICML}, 2020.

\bibitem{AnchorDETR}
Yingming Wang, Xiangyu Zhang, Tong Yang, and Jian Sun.
\newblock {Anchor DETR}: Query design for {Transformer}-based detector.
\newblock In {\em AAAI}, 2022.

\bibitem{MPSR}
Jiaxi Wu, Songtao Liu, Di Huang, and Yunhong Wang.
\newblock Multi-scale positive sample refinement for few-shot object detection.
\newblock In {\em ECCV}, 2020.

\bibitem{FSDetView}
Yang Xiao and Renaud Marlet.
\newblock Few-shot object detection and viewpoint estimation for objects in the
  wild.
\newblock In {\em ECCV}, 2020.

\bibitem{QueryDet}
Chenhongyi Yang, Zehao Huang, and Naiyan Wang.
\newblock {QueryDet}: Cascaded sparse query for accelerating high-resolution
  small object detection.
\newblock In {\em CVPR}, 2022.

\bibitem{reppoints}
Ze Yang, Shaohui Liu, Han Hu, Liwei Wang, and Stephen Lin.
\newblock {RepPoints}: Point set representation for object detection.
\newblock In {\em ICCV}, 2019.

\bibitem{EfficientDETR}
Zhuyu Yao, Jiangbo Ai, Boxun Li, and Chi Zhang.
\newblock Efficient {DETR}: improving end-to-end object detector with dense
  prior.
\newblock {\em arXiv preprint arXiv:2104.01318}, 2021.

\bibitem{PNPDet}
Gongjie Zhang, Kaiwen Cui, Rongliang Wu, Shijian Lu, and Yonghong Tian.
\newblock {PNPDet}: Efficient few-shot detection without forgetting via
  plug-and-play sub-networks.
\newblock In {\em WACV}, 2021.

\bibitem{CADNet}
Gongjie Zhang, Shijian Lu, and Wei Zhang.
\newblock {CAD-Net}: A context-aware detection network for objects in remote
  sensing imagery.
\newblock {\em IEEE Transactions on Geoscience and Remote Sensing},
  57(12):10015--10024, 2019.

\bibitem{Meta-DETR-firstversion}
Gongjie Zhang, Zhipeng Luo, Kaiwen Cui, and Shijian Lu.
\newblock {Meta-DETR}: Few-shot object detection via unified image-level
  meta-learning.
\newblock {\em arXiv preprint arXiv:2103.11731v1}, 2021.

\bibitem{Meta-DETR}
Gongjie Zhang, Zhipeng Luo, Kaiwen Cui, Shijian Lu, and Eric~P. Xing.
\newblock {Meta-DETR}: Image-level few-shot detection with inter-class
  correlation exploitation.
\newblock {\em IEEE Transactions on Pattern Analysis and Machine Intelligence},
  2022.

\bibitem{SAM-DETR-PlusPlus}
Gongjie Zhang, Zhipeng Luo, Jiaxing Huang, Shijian Lu, and Eric~P Xing.
\newblock Semantic-aligned matching for enhanced {DETR} convergence and
  multi-scale feature fusion.
\newblock {\em arXiv preprint arXiv:2207.14172}, 2022.

\bibitem{SAM-DETR}
Gongjie Zhang, Zhipeng Luo, Yingchen Yu, Kaiwen Cui, and Shijian Lu.
\newblock Accelerating {DETR} convergence via semantic-aligned matching.
\newblock In {\em CVPR}, 2022.

\bibitem{DINO}
Hao Zhang, Feng Li, Shilong Liu, Lei Zhang, Hang Su, Jun Zhu, Lionel~M Ni, and
  Heung-Yeung Shum.
\newblock {DINO}: {DETR} with improved denoising anchor boxes for end-to-end
  object detection.
\newblock In {\em ICLR}, 2023.

\bibitem{DA-DETR}
Jingyi Zhang, Jiaxing Huang, Zhipeng Luo, Gongjie Zhang, Xiaoqin Zhang, and
  Shijian Lu.
\newblock {DA-DETR}: {Domain Adaptive Detection Transformer} with information
  fusion.
\newblock In {\em CVPR}, 2023.

\bibitem{graphfpn}
Gangming Zhao, Weifeng Ge, and Yizhou Yu.
\newblock {GraphFPN}: Graph feature pyramid network for object detection.
\newblock In {\em ICCV}, 2021.

\bibitem{m2det}
Qijie Zhao, Tao Sheng, Yongtao Wang, Zhi Tang, Ying Chen, Ling Cai, and Haibin
  Ling.
\newblock {M2Det}: A single-shot object detector based on multi-level feature
  pyramid network.
\newblock In {\em AAAI}, 2019.

\bibitem{ExtremeNet}
Xingyi Zhou, Jiacheng Zhuo, and Philipp Kr{\"a}henb{\"u}hl.
\newblock Bottom-up object detection by grouping extreme and center points.
\newblock In {\em CVPR}, 2019.

\bibitem{DeformableDETR}
Xizhou Zhu, Weijie Su, Lewei Lu, Bin Li, Xiaogang Wang, and Jifeng Dai.
\newblock {Deformable DETR}: Deformable transformers for end-to-end object
  detection.
\newblock In {\em ICLR}, 2021.

\end{thebibliography}
}

\clearpage

\appendix

\section{Technical Appendix}    \label{sec:imfa_appendix}

This section provides more details of our proposed method and its experimental results, which are omitted in the main paper due to space limitation. 

\subsection{Training Objective of Iterative Multi-scale Feature Aggregation (IMFA)}

As described in Section~\textcolor{red}{4}, all additional operations introduced by IMFA are fully differentiable, including the selection of top-K prior detection predictions, sparse feature sampling via bilinear interpolation, adaptive scale selection, Dynamic FFN, and iterative feature aggregation. Thus, the proposed IMFA can be trained end-to-end on top of the corresponding baselines~\cite{DETR,ConditionalDETR,AnchorDETR,DABDETR}.

Besides, IMFA requires no additional training objectives. In other words, IMFA is trained purely with the supervision signals of the corresponding baselines' detection-related losses.

\subsection{Additional Experiment Results}

Table~\textcolor{red}{3} in our manuscript has already demonstrated that our proposed IMFA can work well with stronger vision Transformer (ViT) backbones~\cite{SwinTransformer}. Here we present more results in Table~\ref{tab:supp_additional_results}. With Swin-Transformer-Tiny (Swin-T)~\cite{SwinTransformer} as the backbone, \,\verb+DAB-DETR-Swin-T++\verb+IMFA+\, significantly outperforms \,\verb+DAB-DETR-R50++\verb+IMFA+ with comparable computational cost, which further demonstrates IMFA's excellent scalability.

\begin{table*}[!hbt]
\vspace{+1.5mm}
\begin{center}
\centering
\setlength{\tabcolsep}{5.0pt}
\resizebox{0.92\textwidth}{!}{
\begin{tabular}[t]{l|cc|ccc|cccccc}

\toprule[1.5pt]
Method & MS & SMS & \#Epochs & \#Params & FLOPs & AP & AP$_{\rm 50}$ & AP$_{\rm 75}$ & AP$_{\rm S}$ & AP$_{\rm M}$ & AP$_{\rm L}$ \\

\midrule[1.0pt]

DAB-DETR-R50~\cite{DABDETR}\,+\,IMFA\,(Ours) & & $\checkmark$ & 50 & 53M & 108G & 45.5 & 65.0 & 49.3 & 27.3 & 48.3 & 61.6 \\

\midrule[0.3pt]

\rowcolor{Gray}  \textbf{DAB-DETR-Swin-T\,\cite{DABDETR}\,+\,IMFA\,(Ours)} & & $\checkmark$ & 50 & 57M & 114G & \textbf{47.0} & \textbf{67.1} & \textbf{50.6} & \textbf{29.5} & \textbf{49.7} & \textbf{63.3} \\

\midrule[0.3pt]

Deformable-DETR-Swin-T~\cite{DeformableDETR} & $\checkmark$ & & 50 & 40M & 180G & 45.7 & 65.3 & 49.9 & 26.9 & 49.4 & 61.2 \\

YOLOS-DeiT-B~\cite{YOLOS} & & & 150 & 127M & 538G & 42.0 & 62.2 & 44.5 & 19.5 & 45.3 & 62.1 \\

ViDT-Swin-T~\cite{ViDT} & $\checkmark$ & & 50 & 38M & 100G & 44.8 & 64.5 & 48.7 & 25.9 & 47.6 & 62.1 \\

\bottomrule[1.5pt]
\end{tabular}}
\end{center}
\vspace{-5.3mm}
\caption{Comparison with state-of-the-art object detectors with ViT backbones on COCO val\,2017. `MS' denotes the use of multi-scale features. `SMS' denotes the use of sparse multi-scale features with our proposed IMFA. `\,$\S$\,' denotes two-stage Transformer-based object detector, with the encoder producing `region proposals' to initialize object queries.
}
\label{tab:supp_additional_results}
\vspace{+1.5mm}
\end{table*}

Besides, we also compare our method with other state-of-the-art Transformer-based object detectors using vision Transformers as backbones. As shown in Table~\ref{tab:supp_additional_results}, our \,\verb+DAB-DETR-Swin-T++\verb+IMFA+\, still achieves the best overall performance.
We notice that ViDT~\cite{ViDT} has a lower FLOPs than ours, because it adopts an `encoder-free neck architecture' based on deformable attention~\cite{DeformableDETR}.

\subsection{Additional Visualization Results}

\begin{figure*}[t!] 
\begin{center}
   \vspace{+2mm}
   \includegraphics[width=0.999\linewidth]{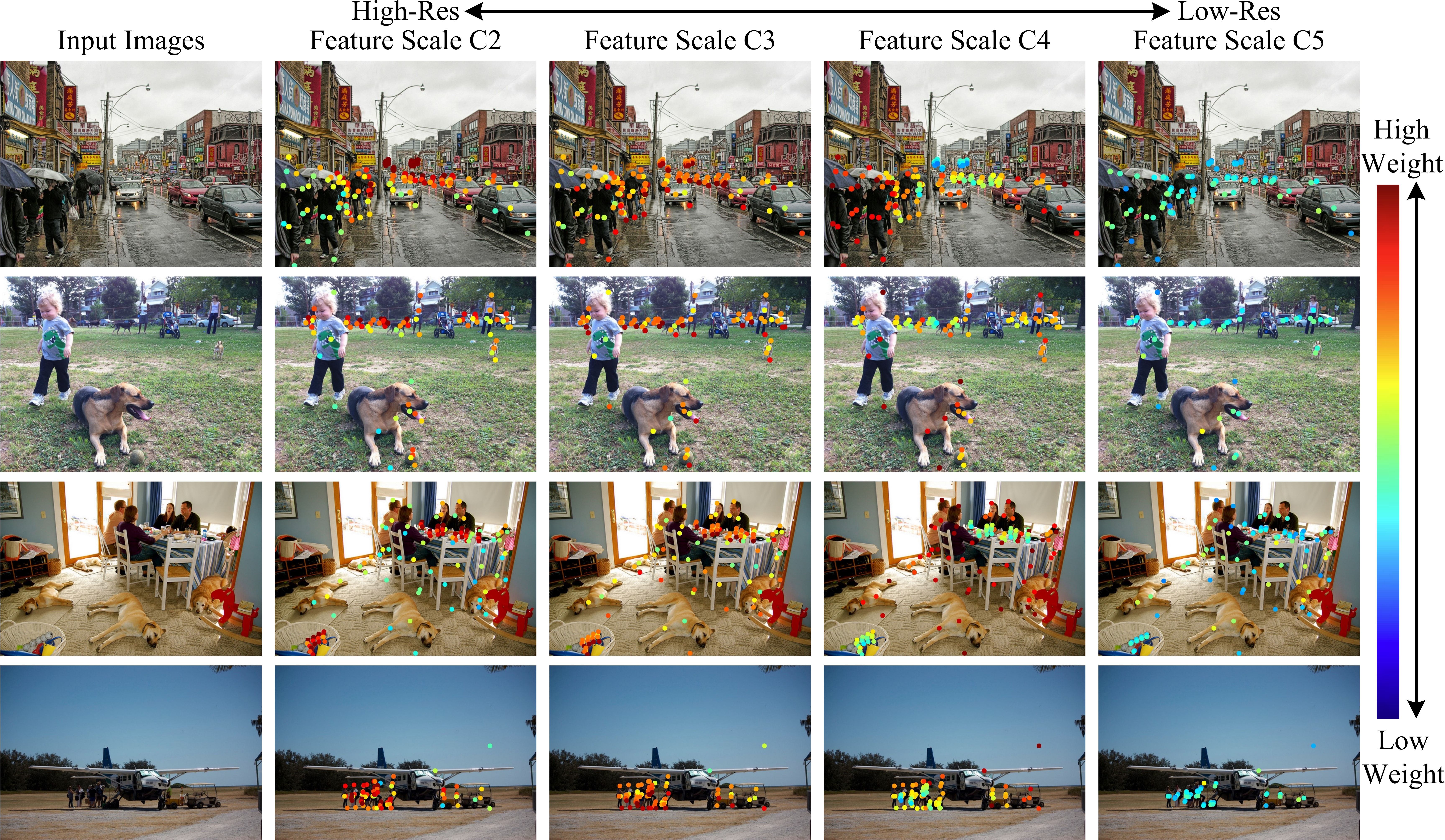}
\end{center}
\vspace{-4.486mm}
   \caption{
   Additional visualization of IMFA's sampling locations and their adaptively selected feature scales. The searched sampling points mostly fall around the objects of interest, many of which are highly representative points with rich semantics, such as objects' extremities. Besides, IMFA adaptively selects appropriate feature scales for each sampling point, generating sparse yet informative scale-adaptive features for refined detection predictions. Best viewed in color.
   }
\label{fig:IMFA_supp_vis}
\vspace{+2mm}
\end{figure*}

For a more comprehensive understanding of the proposed IMFA, we provide more visualization results on IMFA's sampling locations and IMFA's adaptively selected feature scales in Fig.~\ref{fig:IMFA_supp_vis}. These visualizations validate the effectiveness of IMFA in searching informative locations and appropriate scales for multi-scale feature sampling, even under very complex scenarios as shown in the first row.
It is noteworthy that the sampling weights for C5 are generally low, even for large objects. This is because C5 has the same feature scale as the encoded image features, and thus IMFA tends to sample multi-scale features from C2-C4 for additional information.

\subsection{Implementation Details for Human Pose Estimation}

Section~\textcolor{red}{5.4} investigates the generality of IMFA across various tasks by integrating it with PRTR (two-stage variant)~\cite{PRTR}. Here we present the implementation details of this integration.

The implementation details align with PRTR~(two-stage variant)~\cite{PRTR}'s implementation. Concretely, we adopt the person detection results fine-tuned on COCO~\cite{MSCOCO} to extract image patches that contain persons. These image patches are resized into a fixed shape of 384x288, then processed by data augmentations including random rotation, random scale, and horizontal flipping, and finally fed into the \verb|PRTR+IMFA| model. We adopt the AdamW~\cite{AdamW} optimizer for training, with the base learning rate for the ResNet backbone~\cite{resnet} as 1e-5 and 1e-4 for the rest, with a weight decay of 1e-4. The total number of training epochs is 200, and the learning rate is halved at the 120th and 140th epoch, respectively. For the Transformer part, the number of encoder and decoder layers are both set to 6. The number of keypoint queries is set to 100. During inference, we adopt the common practice of flip-test~\cite{PRTR} and compute the keypoint coordinates by averaging the outputs of the original and flipped person image patches.

\subsection{Further Discussions}

\noindent
\textbf{Our differences with multi-scale feature fusion.\;\;}
Compared with existing multi-scale methods (e.g., FPN, DLA, Amulet, Deformable DETR, SMCA-DETR, etc.), the way we utilize multi-scale features is significantly different. Specifically, most existing methods use all the feature tokens from multi-scale features (typically 20x$\sim$80x feature tokens compared to single-scale), whereas IMFA only adds less than 1x multi-scale feature tokens by aggregating multi-scale features from just a few informative keypoints. This is the key reason that IMFA can serve as a generic paradigm for efficient exploitation of multi-scale features in Transformer-based detectors. Our experiments show that, at very slight computational costs, IMFA is able to boost detection performance by large margins for multiple Transformer-based detectors.

It is noteworthy that Deformable DETR~\cite{DeformableDETR} also adopts sparse multi-scale feature computation. However, Deformable DETR still stores and uses all multi-scale feature tokens, which is different from IMFA. IMFA does not need to compute and store dense and high-resolution multi-scale features and is more efficient. Thus, IMFA introduces significantly smaller computational costs in processing multi-scale features.

\vspace{+1.25mm}
\noindent
\textbf{Our relation to guided refinement.\;\;}
Guided refinement typically refers to the recursive update of predictions based on previous predictions. A typical example is Cascade R-CNN~\cite{CascadeRCNN}. Our proposed IMFA falls under the umbrella of guided refinement. However, IMFA's guided refinement is inherited from its baseline methods (e.g., DETR, Conditional DETR, Anchor DETR, DAB-DETR, etc.) that involve multiple decoder layers as refinement stages. We highlight that IMFA does not introduce any additional refinement stages, and neither do we claim IMFA's guided refinement as a novelty or contribution. Our major contribution is that, on top of Transformer-based detectors' guided refinement patterns, we propose IMFA that efficiently and adaptively incorporates new information (sparse multi-scale features) at each detection stage to achieve superior detection performance at slight computational costs.

\end{document}